\documentstyle[psfig]{article}
\newcounter{HoldEx}                  % To hold previous Example number
\newcounter{SaveEx}

\newtheorem{Exmpl}{Example}

\newcommand{\startx} % the start of an example, sets face to \rm and
   {\par\noindent
    \begin{minipage}[t]{5.5in}
    \vspace*{0.5ex}
    \begin{Exmpl}
    \rm \ \\
    \makebox[.25in]{}
    \begin{minipage}[t]{5in}}

\newcommand{\stopx}  % end an example
    {\end{minipage}
     \end{Exmpl}
     \vspace*{0.5ex}
     \end{minipage}\\}

\newcommand{\this}{{\em this}\ }
\newcommand{\that}{{\em that}\ }

\makeatletter
\newcommand{\silentfootnote}[1]
   {\gdef\@thefnmark{~}\@footnotetext{#1}}
\makeatother

\title{Structure and Ostension in the Interpretation\\
of Discourse Deixis}

\author{Bonnie Lynn Webber \\
Department of Computer and Information Science \\
University of Pennsylvania \\
Philadelphia PA 19104-6389\thanks{This work was partially 
supported by ARO grant DAA29-884-9-0027, NSF grant MCS-8219116-CER and DARPA 
grant N00014-85K-0018 to the University of Pennsylvania. I would
like to thank Mark Steedman,
James Allen, Becky Passonneau, Ethel Schuster, Craige Roberts, Jerry  Hobbs,
Annie Zaenen, Mark Gawron, John Nerbonne, Steve Levinson,
Barbara Di Eugenio, Robin Cohen and William Marslen-Wilson
their helpful comments. A preliminary version of this paper was 
presented at the 26th Annual Meeting of the Association for Computational 
Linguistics, Buffalo NY, June 1988.}} 
\date{}

\begin{document}           % End of preamble and beginning of text.

\maketitle                 % Produces the title.

\begin{abstract}

This paper examines demonstrative pronouns used as {\em deictics}
to refer to the interpretation of one or more clauses. Although this
usage is frowned upon in style manuals (for example Strunk and White (1959)
state that ``{\em This}. The pronoun {\em this}, referring to the complete
sense of a preceding sentence or clause, cannot always carry the load and
so may produce an imprecise statement.''), 
it is nevertheless very common in written text.
Handling this usage poses a problem for Natural Language Understanding
systems. The solution I propose is based on distinguishing
between what can be {\em  pointed to} and what can be {\em
referred to} by virtue of pointing. I argue that a restricted set of
{\em discourse segments} yield what such demonstrative pronouns can point
to and a restricted set of what
Nunberg (1979) has called {\em referring functions} yield what they can
refer to by virtue of that pointing.
\end{abstract}

\section{Introduction}

\subsection{The Phenomenon}
\label{Intro}

This paper sets out to explicate a use of demonstrative pronouns illustrated in
the following examples:

\startx
It's always been presumed that when the glaciers receded, the 
area got very hot. The Folsum men couldn't  adapt, and they died out. 
{\em That}'s what is supposed to have happened. It's the textbook 
dogma. But it's  wrong.
\stopx
\startx
Using microscopes and lasers and ultrasound, he removes tumors that
are intertwined with children's brain stems and spinal cords. There is
only the most minute visual difference between the tumors and normal tissue.
Operations can last 12 hours or more. The tiniest slip can kill, paralyze or
leave a child mentally retarded.

{\em This} is the easy part of his job. \hspace*{2em}({\em New York Times},
11 August 1990)
\stopx
In these examples, a demonstrative pronoun is used
to refer to something other than the referent of a previous noun phrase.
It is not, however, referring to a previous {\em section of text}
-- what Lyons (1977, p.668) has called {\em pure textual deixis} -- as in,

\startx
A: The combination is 7-4-1-5.\\
B: I'm sorry I didn't hear you. Could you repeat {\em that}?
\stopx
In example~1, the referent of {\em that} must be something that can
happen. In example~2, it must be something that can be part of a job.
Sections of text, i.e., strings of words, can be neither. On the other hand,
sections of text can have such events or actions associated with them.
My goal in this paper is thus straightforward, i.e. to show how sections
of text can provide referents for demonstrative pronouns. 

The paper proceeds as follows:  First I discuss terminology and assumptions. I
then try to show (Section~\ref{DSegs}) that it is not any {\em arbitrary
sequence of clauses} in a text that can provide referents for demonstrative
pronouns, but rather, only those that correspond to what have been called {\em
discourse segments} in many current discourse theories. I then try to show
that,
at any point in the text, only certain discourse segments can yield referents
for demonstrative pronouns -- in particular, only those segments whose
{\em contribution to the discourse model} is currently {\em in focus}.

To make this sense of focus precise, I present in Section~\ref{ITC}
a simple incremental tree
construction algorithm which is meant to serve as a {\em formal analogue} for
text processing. This algorithm specifies precisely (1) the positions at which
new nodes can be inserted into a tree and (2) the particular insertion operations
that can be used. I then associate with these positions (the current {\em right
frontier} of the tree) the set of discourse segments whose contribution to the
discourse model is in focus. I claim that it is just these discourse
segments that can yield referents for demonstrative pronouns. In
support of this claim, I provide empirical evidence (Section~\ref{IDP}) that
segments corresponding to nodes no longer on the {\em right frontier} can no
longer provide such referents.

Thirdly, I argue against the view that it is the structure of the
{\em world} rather than how a speaker chooses to describe it that is
the primary constraint on what deictic pronouns can refer to in discourse.
I conclude in Section~\ref{Bronto} with a descripiton of the indirect process
by which I believe discourse segments yield referents for demonstrative
pronouns through their contribution to the discourse model. This process
draws on Nunberg's exploration into polysemy and demonstrative reference
(Nunberg, 1979) and current theories of what has been called {\em natural
language metaphysics}  (Bach, 1989).

\subsection{A Name for the Phenomenon}

Phenomena need names. Lyons, recognizing the difference between
{\em pure textual deixis} (see above) and the phenomenon
under discussion here, called this phenomenon {\em impure textual deixis}. It
was ``deixis'' because
\begin{quote}
The term `deixis' (which comes from a Greek word meaning ``pointing'' or
``indicating'') is now used in linguistics to refer to the function of
personal and demonstrative pronouns, of tense and of a variety of other
grammatical and lexical features which relate utterances to the spatio-temporal
co-ordinates of the act  of utterance. (Lyons, 1977, p.636)
\end{quote}
It was ``textual'' because it had to do with the utterance itself, and
``impure'' because what was being indicated was not the utterance as a
thing but what it expressed.

I think a better name is called for. The terms {\em reference to events} and
{\em reference to propositions} used by B\"{a}uerle (1989) reflect
the semantic {\em sort} of the thing referred to. My problem with these
terms are that (1) events and propositions are only two of the many sorts of
things that demonstrative pronouns can refer to in discourse and (2) the
separate names may be seen as implying that there are separate processes
involved in reference to events and in reference to propositions. I do not
believe this. I try to show here that only a single process need be involved.
Another possible name is {\em discourse deixis}, previously
used by Lakoff (1974). This is the name I favour: it labels the
phenomenon as an instance of deixis, and grounds the {\em source} of its
referents in the discourse.

As for a name for the referring phrases, when Lyons begins his discussion
of \this and {\em that}, he uses the locution {\em the English demonstratives
`this' and `that', used as deictics} (Lyons, 1977, p.~655)
acknowledging the fact that demonstrative pronouns are used for other
functions as well. Shortly thereafter, he contracts this locution to
{\em deictic pronoun} --
\begin{quote}
In so far as the very fact of pointing to something commits the person who
is pointing to a belief in the existence of what he is pointing at,
the use of a deictic pronoun carries with it
the implication or presupposition of existence. (Lyons, 1977, p. 656)
\end{quote}
One reason for promoting this phrase is that it turns out that in several
languages, including
Italian and even English, zero-pronouns (\O) can be used in the same
way as demonstrative pronouns. (See Di Eugenio (1989) for a discussion of
discourse deixis in Italian.) In English, this occurs in instructions, where
it is common to find ellipsed direct-objects (Sadock, 1974). However, the
referent of this ellipsis need not be a physical object. Consider example~4
drawn from a Frigidaire assembly and repair manual:

\startx
Check the door seal by closing the door on a 1'' wide strip of paper. A slight
drag should be felt when the paper is pulled from between the gasket and the
cabinet. Repeat \O\  around all four sides of the door.
\stopx
Here the zero-pronoun refers to the process described in the preceding
section of text.\footnote{Dick Oehrle has bointed out that these instructions
literally specify five tests: it is only three additional repetitions that
are needed. But how much in life would get done if people insisted on
following instructions literally?} This reference would ordinarily be
achieved explicitly using
a demonstrative  -- ``Repeat this (or this process) around all four sides
of the door.'' Thus the term
{\em deictic pronoun} can serve to denote any pronoun (zero-pronoun,
demonstrative pronoun, or even personal pronoun) that serves this same
function.

\subsection{Terminology and Other Assumptions}
\label{terminology}

I assume that in processing a text, a listener is constructing a {\em model}
that supports it and that evolves with the text. Lyons (1977, p.670) has
called this model the {\em universe-of-discourse}. I and others have called it
a {\em discourse model}. A discourse model is clearly a mental construct, but
as long as the participants in a discourse believe they understand
one another, they will assume that their models are consistent. Discourse
models may correspond to particulars of the real world, if the real world is
the subject of the discourse, but they need not. (This is not to say that
people do not use their beliefs about the real world and how it works, in
interpreting texts and constructing discourse models. It just says that the
particulars of the two need not be the same. The only thing
essential for successful discourse is that the participants believe their
models are consistent with one anothers'.)

A discourse model contains {\em entities} to which are ascribed the
properties and relationships predicated of them in the text. (Bill Woods
once referred to them as ``conceptual coathooks''.) More sober names that have
been used include {\em discourse referents} (Kartunnen, 1976), {\em reference
markers} (Kamp, 1981), and {\em file cards} (Heim, 1983). I have called them
{\em discourse entities} (Webber 1979, 1982), and will continue to do so in
this paper. Discourse entities ground {\em referring phrases}. They may or
may not correspond to particular entities in the real world.

As in many recent theories of discourse, including {\em Discourse
Representation
Theory} ({\em DRT}) (B\"{a}uerle,1989; Kamp, 1981; Roberts, 1989), Grosz and
Sidner's theory of discourse (1986) and that of Polanyi (1986), I assume that
a discourse model is {\em structured} into regions in a way that reflects the
recent structure
of the evolving text. Utterances seen as part of the same segment of text (for
any of the reasons noted in Section~\ref{DSegs}) will be taken by the listener
to contribute to the same region of the model.\footnote{It is not necessary
to suppose that a particular substructure {\em persists} indefinitely. Later
(Section~\ref{DSITC}), I will note how long it is needed in order for the
current approach to work.} Likewise, clauses in the text that are
understood as contributing to the same region of the model will
be seen as part of the same segment of text. Regions of the model may be
embedded in other regions, reflecting substructures in the text. In DRT, these
regions correspond to contexts. As in recent work in DRT (Asher, 1987;
B\"{a}uerle, 1989), I assume that each context has a discourse entity that
``stands proxy'' for its propositional content. This discourse entity will
come into play in Section~\ref{Bronto}, when I explain how discourse
segments can indirectly yield referents for deictic pronouns through their
associated regions in the discourse model.

Like Nunberg (1979) and others, I assume an approach to ostensive acts (such
as in the use of deictic pronouns) that
distinguishes what is pointed to (the {\em demonstratum}) and what is
referred to (the {\em referent}). These two entities may be the same, but
they do not have to be. What is of interest is the link between them, which
Nunberg (1979) has called a {\em referring function}. Referring functions
apply to demonstrata to produce referents.
In the approach that I present in
Section~\ref{Bronto}, referring functions apply to discourse entity
``proxies'' for regions of the discourse model corresponding to discourse
segments, thereby yielding referents for deictic pronouns. The range of
possible
referents follows directly from the range of {\em referring functions} possible
in the given circumstance. In example~5, at least four are possible:

\startx
Hey, management has reconsidered its position. They've promoted Fred to second
vice president.\\
(a.) {\em That's} a lie. \\
(b). {\em That's} false. \\
(c.) {\em That's} a funny way to describe the situation.\\
(d.) When did \that happen?
\stopx
In (a), the referenced interpretation is the specific speech act (only speech
acts can be lies); in (b), the proposition conveyed by the segment; in (c) the
description expressed by the segment; and in (d), the particular event denoted
by the segment. Because in the rich ontology advocated by Bach (1990) and
others, all of these are sorts of individuals, I take it that the act of
{\em ostension}
performed by deictic pronouns in discourse can add new individuals
(discourse entities) into the model that were not present prior to the
ostensive act. As such, ostension can have the same effect as what has been
called {\em accommodation} (Lewis, 1979).

That is, in the simplest case, a referring phrase will be taken to
refer to an entity already in the discourse model. However, it can also
cause the listener to add a new entity to the model to which he or she
can ascribe the indicated properties or can set in the indicated relations.
The addition of a new entity in response to an indefinite noun phrase is a
common case. Where new entities are added in response to definite noun
phrases, the process has been
called {\em accommodation} because the use of a singular definite is felt to
presuppose that there is {\em already} an unique entity in the context with the
given description that will allow a truth value to be assigned to the
utterance. For example:

\startx
I walked up to the first house on my list. I noticed that {\em the side door}
was wide open.
\stopx
Houses do not necessarily have side doors. However, in response to ``the
side door'', the listener accommodates a new entity in his/her discourse model
corresponding to the mentioned side door of the house in question
(``mentioned'', because the text is consistent with there being more than one
side door).  Deictic pronouns can have the same effect as such definite
noun phrases.

Finally, I assume that one can identify (at least, {\em a posteriori}) what
a speaker and listener are attending to -- what is immediately salient --
at any point in the discourse. Items so identified are considered {\em in
focus}, although other terms have been used to distinguish particular theories.
Notions of {\em focus} have been proposed, at least in part, to account
for patterns of  ``efficient'' concept
verbalization, e.g. when  the pronunciation of concept descriptions
can be attenuated, when  concepts can be specified using explicit pronouns or
zero-anaphors,  when an unmodified definite noun phrase can be used to refer to
a  concept, when particular intonation structures and/or marked 
syntactic constructions like clefts are appropriate, etc. Among the {\em
sorts} of things that are felt to be able to have sort of ``focussed''
status are discourse entities (Grosz et al., 1983; Sidner, 1982), open
propositions (Prince, 1986; Steedman, 1990; Wilson \& Sperber, 1979) and
``focus spaces'' (Grosz \& Sidner, 1986; Reichman, 1985).

I will use the notion of {\em focus} to distinguish those regions of the
discourse model that, at a given point in the discourse, can yield referents
for deictic pronouns. The main feature of focus that is relevant here
is that it changes in fairly predictable ways, as
each new clause is processed.

While I will discuss focus with respect to which discourse segments can yield
referents for deictic pronouns, when there is more than one, as in
the following minimal pair, I will have nothing to say here about how the
choice between them is made.

\startx
{\bf a.} Segal, however, had his own problems with women: he had been trying
to keep his marriage of seven years from falling apart; when {\em that} became
impossible \ldots \\

{\bf b.} Segal, however, had his own problems with women: he had been trying
to keep his marriage of seven years from falling apart; when {\em that} became
inevitable \ldots
\stopx
In Example~\theExmpl a. , it is the region associated with ``(Segal's) keeping
his marriage of seven years from falling apart''
that yields a referent for {\em that}, while in Example~\theExmpl b., it is
``his marriage of seven years falling apart'' that does so.

\subsection{For the Record}

\begin{figure}
\centerline{\psfig{figure=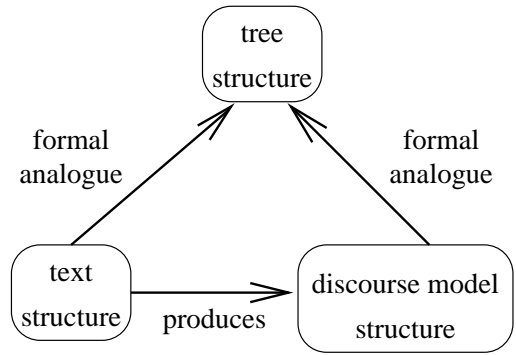}}
\caption{Three Relevant Structures}
\label{THREE}
\end{figure}
It is important for the reader to be clear that there are {\em three}
structures under discussion here (see Figure~1): the structure of the text
(in terms of discourse segments), the structure of the discourse model (in
terms of regions), and a tree structure which, by
virtue of its associated insertion procedure, serves as a {\em formal 
record} of the {\em process} by which the discourse model grows
and the structure changes, in response to the text and its perceived
structure. The value of having such a formal analogue lies in being able to
make
use of its properties, including how it can legally change over time, to pin
things down and avoid hand-waving. (In Section~\ref{DSITC}, I discuss
what I believe is the close relationship between these three structures
and those proposed by Grosz and Sidner.)

I should also note that this is not the first attempt to provide an account
of demonstrative pronouns in discourse (nor do I expect it will be the last).
Both Linde (1979) and Sidner (1982) tried to extend theories of anaphoric
pronoun interpretation to account for demonstrative pronouns as well. Both
these accounts are in terms of the {\em focus} status of different objects
and actions mentioned in the discourse. More recently, though, Passonneau
(1989) analysed 678 instances of {\em it} and {\em that} found in
conversational interactions, where these pronouns are interpreted as
referring to entities introduced  into the discourse model by noun phrases
and other sentential constituents. She found two independent factors that
strongly predicted whether subsequent reference would be via {\em it} or via 
{\em that}: one she called {\em persistence of grammatical subject}, the 
other, {\em persistence of grammatical form}. By the first factor, if
both antecedent and pronoun were {\em subjects} of their 
respective clauses: if so, the pronoun {\em it} was strongly favored.
By the second factor, if the antecedent noun phrase was other than a
pronoun or a canonical noun phrase headed by a noun,
then the pronoun \that was strongly favored. Unlike the current study,
Passonneau's analysis was limited to pronouns whose antecedents were
constituents within single sentences.

There are also at least two attempts to extend Kamp's (1981)
Discourse Representation Theory to handle reference to events and
propositions (Asher, forthcoming; B\"{a}uerle, 1988), but as I will
try to show, it is more than just events and propositions that deictic
pronouns can refer to in discouse, so that a richer interpretive process
is called for.

\section{Discourse Segments}

\subsection{Background}
\label{DSegs}

I hope that a simple example will convince the reader that not every previous
sequence of clauses in a text can yield referents for deictic
reference. Consider (8):

\startx
a. For his part in their joint project, John built a two-armed robot.\\
b. He had learned about robotics in CSE391.\\
c. For her part, Mary taught it to play the saxophone.
\stopx
It is simple to come up with an subsequent utterance in which the referent
of \that derives from just the previous clause (8c), for example:
\begin{quote}
d. {\em That} took her six months.
\end{quote}
It is also easy to come up with a subsequent utterance in which the
referent of \that derives from the previous three clauses (8a-c), for example:
\begin{quote}
d'. That earned them both As.
\end{quote}
However, it does not seem possible to come up with a subsequent utterance in
which the referent of \that derives from just the previous two clauses (8b
and c). This fact seems to follow from the fact that they are not
interpreted together as a unit, independent of 8a.

If not every sequence of clauses in a text has an interpretation accessible to
deictic reference, which ones do? 
There is a widely held view that discourses are formed of smaller 
sequences of related clauses or sentences called {\em discourse 
segments}, although as James Allen (1987, pp.398-9) has noted:
\begin{quote}
\ldots there is little consensus on what the segments of a particular
discourse should be or how segmentation could  be accomplished. 
One reason for
this lack of consensus is that there is no precise definition of what a
segment is  beyond the intuition that certain sentences naturally 
group together.
\end{quote}
Among the bases that have proposed for grouping utterances into segments 
are: common conversational role (Fox, 1987; Hinds, 1979), common
{\em discourse 
purpose} with respect to a speaker's plans (Grosz \& Sidner, 1986); common 
meaning (Hobbs, Stickel, Martin \& Edwards, 1988); common perspective in
describing a single event
(Nakhimovsky, 1988), and common modality (Roberts, 1989).
As for what constitutes a minimal discourse segment, theories differ. Hobbs
et al. (1988) take it to be a sentence, and Polanyi (1986), a clause.  Grosz
and Sidner (1986) seem to take a sentence as the minimal segment needed to
express a single purpose, but, unlike Hobbs et al. and Polanyi, do not
assume that every sentence constitutes a distinct discourse segment.

Characterizing what is involved in recognizing that utterances share a
common subject, viewpoint, modality and/or purpose (and doing so in a manner
amenable to computer implementation) is both an unsolved problem and an active
area of research. To date, particular lexical
and syntactic cues have been identified as signalling segmental changes
(Cohen,1987; Grosz and Sidner,1986; Nakhimovsky, 1988; Reichman, 1985), as have
specific intonational changes at segment boundaries (Hirschberg and Litman,
1987).

That is not the whole story, however. Also needed is a characterization of how
listeners bring their beliefs about the world and about speakers' intentions to
bear on recognizing common subject, purpose, etc., as this cannot be done
on the basis of surface and syntactic cues alone. Here there are fewer
research results to date.  For this paper, I will have to assume,
like Cohen (1987), the existence
of an oracle that can decide with which, if any, existing segment the next
utterance in a text shares a common subject, viewpoint, etc. If doing so leads
the reader to abandon the paper at this point, so be it. My feeling is that
existing evidence that discourse segments play a role in text understanding
makes it worthwhile to continue, in parallel, efforts at characterizing
recognition procedures and efforts like the present one that assume such
recognition.

What roles have discourse segments been seen to play in discourse
understanding? One early computational reason for appealing to the notion
of {\em discourse segment} was as a {\em domain of locality} for definite
noun phrases (Grosz, 1977, 1981), to account in part for the fact that the
same definite noun
phrase may refer to different discourse entities at different points in the
discourse. The claim is that a definite noun phrase would be interpreted as
referring to an entity mentioned in the same segment rather than one mentioned
anywhere else, even if the latter were mentioned more recently. For example,
consider the following sequence uttered by a single speaker:

\startx

a. Do you think I can borrow your tent?\\
b. The one I took on my last hike leaked,\\
c. and I haven't had time to replace it.\\
d. I would of course have {\em the tent} cleaned before
returning it to you.
\stopx
By {\em the tent} in (\theExmpl d), the speaker is referring to the one
she has requested in
(\theExmpl a), and not the leaky one she has mentioned in (\theExmpl b,c).
This can be explained in terms of discourse segments, by saying that
clauses (\theExmpl b,c) make up a segment embedded in the larger one (\theExmpl
a-d). As a whole, the segment constitutes a request, with (\theExmpl b,c)
functioning as its explanation. Because (\theExmpl d) is outside the embedded
segment, its object noun phrase {\em the tent} would be interpreted as
referring to the one mentioned in the segment that it belongs to (i.e., the
outer one).

Another use to which the notion of discourse segment has been put is in the
interpretation of tensed clauses in narrative text (Nakhimovsky, 1988; Webber,
1988). For example, the unmarked interpretation of a sequence of simple past
tense clauses in English narrative is that the events described happen at the
same time or in temporal sequence. But this is not always the case, as in
the following example.

\startx
a. I was over at Mary's house yesterday.\\
b. She told me about her brother Harold.\\
c. He went to Alaska with two friends.\\
d. Together they climbed Mt. McKinley.\\
e. She asked me whether I was interested in going some time.
\stopx
Clearly the event described in clauses (10c,d) happened before the event
described by (10a,b,e), even though all the clauses use the same simple past
tense. Postulating an embedded segment and a temporal focus that can reset at
segment boundaries (Webber, 1988) allows one to retain a notion of simple
simultaneity or temporal progression in the unmarked case.

\subsection{Relations between Discourse Segments}

Often (as above) {\em discourse segments} are taken to be recursive
structures, such that either a discourse segment is a minimal
segment or it comprises a sequence of embedded discourse
segments.\footnote{As Passonneau has 
pointed out (personal correspondence), this ignores the possibility
of interpreting a stretch of text as belonging to
two adjacent segments in a sequence, serving essentially as a
transition between them. Including this possibility complicates
what it would mean to have a {\em sequence} of discourse segments,
but would not alter the recursive nature of the definition itself.}
As so defined, the recursive structures of interest are
{\em trees}. This does not mean that a discourse corresponds to a {\em single
tree}, just that there  may be parts of the discourse that
evince an embedding structure, and that this structure has interesting
properties.

Now, if a tree structure is to represent the relationship among
(certain) segments though, then so must its two basic structuring relations --
{\em parent-of} and {\em right-sibling-of}. For example, in Robin Cohen's
work on the structure of argumentative discourse (Cohen, 1983; 1987),
{\em parent-of}
means that the claim made by the child provides evidence for the
claim made by the parent. {\em Right-sibling-of} corresponds
to the linear order of claims that provide evidence for the same
conclusion. Cohen's goal is to understand how structured arguments
are transmitted through a linear sequence of clauses. She presents
three common {\em transmission forms} that enable
minimal effort reconstruction of the structure underlying
an argument: pre-order, post-order and a mixed pre- and
post-order. These tramission forms require minimal
effort because of the severe restrictions they place on what an incoming
clause can stand in a parent/child or sibling relation to.
Cohen shows how ``clue words'' 
can be used to provide enough information to enable departures 
from these expected transmission forms and still produce 
comprehensible arguments.

In Scha and Polanyi's proposal (1988) for a semi-deterministic, on-line
procedure for building a structural description of
an unfolding discourse, they take the nodes of
discourse structure trees to be any of a variety of types of {\em discourse
constituent units} (DCUs). DCUs differ from one another in two ways:
(a) how they derive their semantic attributes from those of their
constituents, and (b) the ``accessibility'' of their constituents
to things like anaphoric reference. (The three types of DCU discussed
in (Scha and Polanyi, 1988) are subordinations, binary coordinations,
and n-ary coordinations, each of which has several subtypes. For
example, lists and narratives are types of n-ary coordinations.) For all these
node types, {\em parent-of} means uniformly that one {\em DCU} is a constituent
of another one. However, the meaning of {\em right-sibling-of} varies,
depending 
on the type of common parent node. For example, {\em right-sibling-of} in a
{\em narrative} n-ary coordination has a temporal aspect to its meaning,
which it doesn't in a {\em list} n-ary coordination.

Grosz and Sidner (1986) take a more abstract criterion for
establishing structural relations in discourse.  The {\em parent-of}
relation they call {\em domination} (DOM), and the {\em sibling-of}
relation, {\em satisfaction-precedes}. They take these relations to
hold between what they call {\em discourse segment purposes} or DSPs,
rather than between discourse segments directly.  A segment's DSP
specifies how it contributes to achieving the overall discourse
purpose. If the DSP of one segment serves to satisfy that of another,
the latter {\em dominates} (or stands in a {\em parent-of} relation
to) the former. If one DSP must be satisfied before another (in
satisfying some larger purpose), than the former {\em
satisfaction-precedes} the latter (or, alternatively, the latter is the
{\em right-sibling-of} the former).
Grosz and Sidner call the resulting hierarchy of DSPs the {\em
Intentional Structure} of a discourse. It is only one of three structures that
they associate with a discourse. Another of the three, {\em
Attentional State}, I will discuss in Section~\ref{DSITC}.

\subsection{Incremental Tree Construction}
\label{ITC}

Having specified the two tree-structuring relations {\em parent-of} and
{\em right-sibling-of}, there is still a variety of
ways that a tree can be grown incrementally from elements added to it in
sequence, depending on the insertion operator(s) used and the
nodes that insertion can apply to. Here I will show how the same initial
tree and input sequence result in two different final trees, by employing
different insertion operations.

First, consider a {\em binary search tree} (BST), i.e. a tree with a maximum
of two branches on each node and the restrictions that (1) the value of any
node on a left subtree is {\em less than} the value of its root and
(2) the value of any node on a right subtree is {\em greater than} the
value of its root. Here there are two simple operators {\em attach as
left daughter}
and {\em attach as right daughter}, where the position at which a new node is
attached depends on how its value compares with that of existing nodes on a
path from root to leaf. These operators can insert new nodes only at
the {\em fringe} of the tree. Thus the root of the tree can never change,
nor can the relation between any existing node and the root. Figure~2
illustrates the incremental change in a BST as two new elements, whose
values are 50 and 25, are added in sequence.
\begin{figure}
\centerline{\psfig{figure=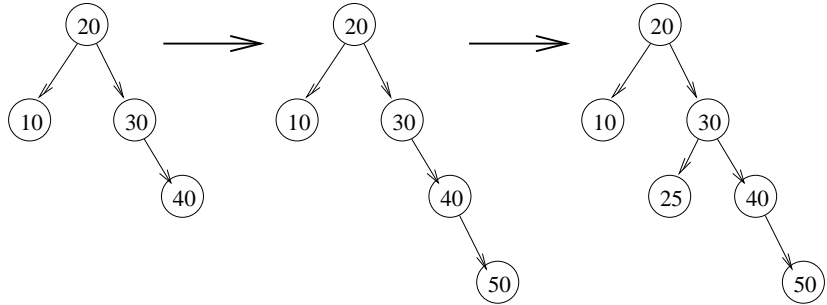}}
\caption{Growing a Binary Search Tree}
\label{BST}
\end{figure}

Now consider another data structure, the {\em AVL tree}. An AVL tree must
satisfy the same constraints as a BST, as well as one additional
constraint: an AVL tree must be kept {\em balanced} around each node.
(A tree is considered balanced if the heights of its left and right
subtree do not differ by more than 1.) Again, new nodes can only be
added at the fringe of the tree, but the insertion operations are more
complex, often leading to a restructuring of the tree (via
rotation) to keep the tree balanced (see Reingold and Hansen, 1983, or any
standard text in data structures, for a more detailed discussion).
\begin{figure}
\centerline{\psfig{figure=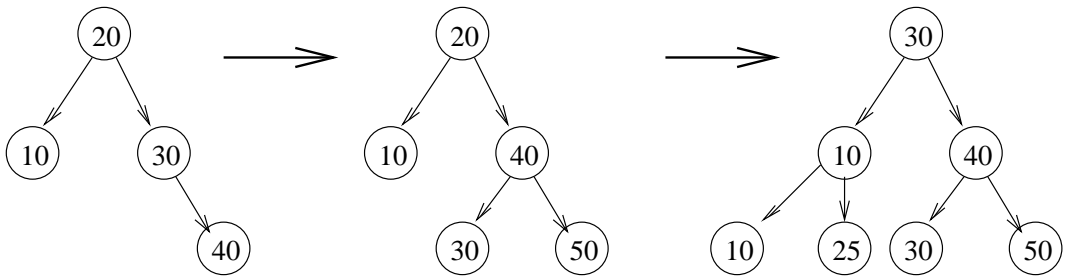}}
\caption{Growing an AVL Tree}
\label{AVL}
\end{figure}
Figure~\ref{AVL} illustrates the incremental change in the same initial
tree -- which also qualifies as an AVL tree -- as the same two elements,
50 and 25, are added in sequence. Notice that the root node of the AVL has
changed by virtue of the insertions, as has the fringe and the right
frontier. (The {\em right frontier} of a tree comprises those nodes along
the path from root to tip defined by the sequence of rightmost daughters,
starting at the root, as shown in Figure~4.)
\begin{figure}
\centerline{\psfig{figure=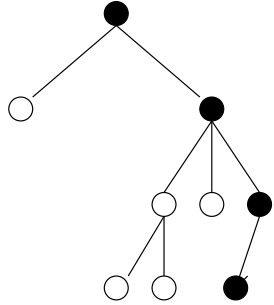}}
\caption{Right frontier}
\end{figure}

No one has claimed that BSTs or AVL trees are relevant to discourse
understanding. However, they can provide a basis for understanding
other procedures that have been proposed for discourse understanding.
For example, Cohen's pre-order transmission form
(Cohen, 1983, 1987) uses a single operation -- {\em attach as rightmost
daughter}. Her post-order transmission format makes use of a different
operation
-- {\em attach as parent} -- which subordinates a node or set of nodes as
daughters of the node corresponding to the new  clause. Her hybrid strategy
makes use of both operators. At any point, there is an identifiable set of
nodes
that these operators can apply to. When an operator applies, it may change the
tree such that different nodes are available as points of attachment. As noted,
Cohen allows for linguistic clues to redirect operators to other nodes in the
tree.

What I want to propose is a somewhat different, incremental procedure that
makes use of two operations -- {\em attach as rightmost  daughter}
(or simply, {\em attach}) and {\em adjoin}. After describing the operators,
where they apply, and how they change the shape of the evolving tree, I will
try to explain in discourse terms what the two operators are meant to
correspond to. I then offer evidence (Section~\ref{IDP}) in support of the
claim that it is regions of the discourse model {\em corresponding to}
nodes on the
right frontier of a tree created by such a procedure that can yield referents
for deictic pronouns.

\begin{figure}
\centerline{\psfig{figure=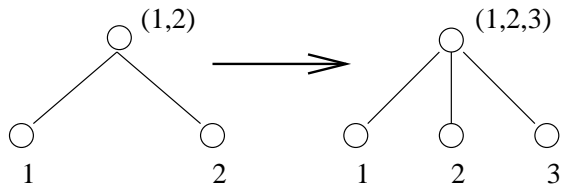}}
\caption{Addition of Nodes by Attachment}
\label{attach}
\end{figure}
First, the operations. {\em Attach} is the simple operation {\em attach as
rightmost daughter} discussed previously. (Because nodes here do not
have explicit values, I will label leaf nodes in order of their incorporation
amd non-terminal nodes, with a list of their children in left-to-right order.)
{\em Adjunction} is a somewhat more complex operation which creates new
non-terminal
nodes rather than just adding leaves to existing non-terminals. When a new node
is {\em adjoined} to one already in the tree, both become children of the same
(new) parent node, with the new one being the right-sibling of the old.
As I will show in the next sectin, adjunction applies slightly differently
to root, non-terminal and leaf nodes. However, it always adds two nodes to a
tree, whereas attachment adds only one.

In the procedure described here, {\em attachment} and {\em adjunction} are
limited to nodes on the {\em right frontier} of the evolving tree
structure. (In Figure~4, nodes on the {\em right frontier} are shaded.)
If this were really an independent procedure I was describing, I would
have to specify how the choice would be made as to (a) {\em where} on
the right frontier to add a new node and (b) {\em how} to do so
(i.e., by {\em attachment} or {\em adjunction}). However, because my purpose
is just to use the procedure to enable the reader to visualize
what happens during text processing to the {\em right frontier} of this
{\em discourse tree}, I will describe these choices in discourse terms.

\subsection{Discourse Segments and Incremental Tree Construction}
\label{DSITC}

I will start by assuming a one-to-one mapping between discourse segments and tree
nodes, with a clause constituting the minimal segment. When the next clause in a
text is taken (for whatever reason mentioned in Section~\ref{DSegs}) to be
included in an existing discourse segment, the corresponding tree operation is
{\em attachment} at the node corresponding to that segment. For example, when
the third clause is processed in (11), it
corresponds to a node being attached to the root of the same tree
(Figure~\ref{attach}).

\startx
1. John eats yoghurt for breakfast,\\
2. Fred eats Cheerios,\\
3. and Mary, muffins.
\stopx
At the level of the discourse model, attachment means that the region of
the discourse model comprising the entities, properties and relations
conveyed by the new clause is included in the same region of the model
as those associated with the rest of the segment. Both the entire region
and the new subregion correspond to nodes on the right frontier of
the current DST. It is these regions -- ones corresponding to nodes
on the right frontier of the current DST -- that will be taken to be {\em in
focus} (Section~\ref{terminology}).

Now for the discourse correlate of {\em adjunction}. When a discourse segment
S$_{i}$ is taken as being directly embedded in another segment S$_{j}$, the
assumption is that the former contributes {\em directly} to the meaning or
purpose of the latter. Suppose clause C comes along, causing the listener to
realize that it is not S$_{i}$ that contributes directly to S$_{j}$, but rather
the combination of S$_{i}$ and C. That is, S$_{i}$ and C form a segment
directly embedded in S$_{j}$.  This is the discourse correlate of {\em
adjunction} to a non-terminal or leaf node. If C is seen as contributing along
with the top-level segment (corresponding to the current root of the tree) to
some more encompassing meaning or purpose, this is the discourse correlate of
{\em adjunction to the root}. The general case of adjunction to the root is
shown in Figure~6.
\begin{figure}
\centerline{\psfig{figure=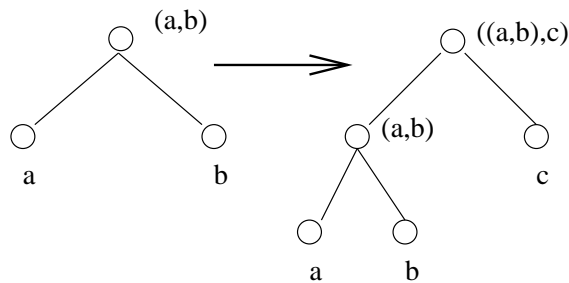}}
\caption{Adjunction to root node}
\end{figure}

Discourse processing that correlates with {\em adjunction to the root} is
actually a common case, as it happens whenever
the meaning or purpose of the second clause in a discourse is taken to
combine with that of the first clause to form a more encompassing meaning or
purpose, such as when the second clause of (11) is processed. This
simple common case of adjunction ot the root is shown in Figure~7.
\begin{figure}
\centerline{\psfig{figure=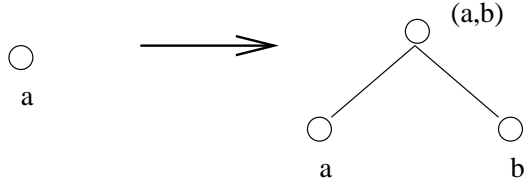}}
\caption{Adjunction to the root of a one-node tree}
\label{OneNode}
\end{figure}

To understand the discourse correlate of {\em adjunction to a
leaf node}, consider the following text at the point
at which the reader has finished processing the clause, ``John hates
snakes''.

\startx
a. John hates snakes.\\
b. His next-door neighbour had kept snakes,\\
c. and he had hated his neighbours.
\stopx
In processing the second clause (\theExmpl b), the reader may decide that it
provides an explanation for the situation described in the first clause and
that the two thereby constitute a segment. (This correlates with the
{\em adjunction to the root} operation described above. It is shown in
step~1 of Figure~8.)
\begin{figure}
\centerline{\psfig{figure=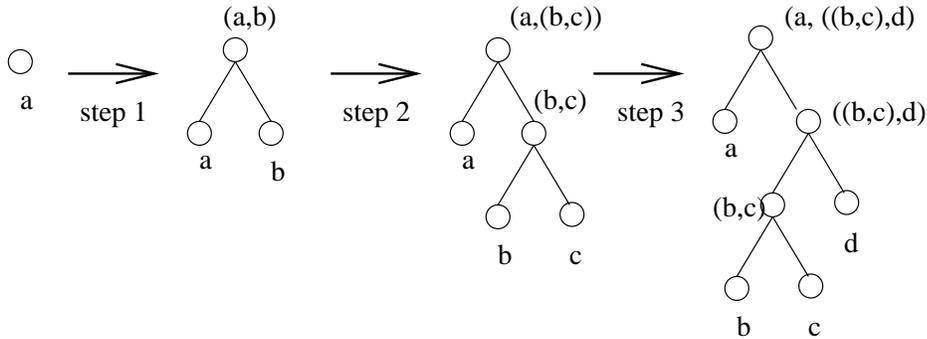}}
\caption{Adjunction operations corresponding to example~12}
\label{SBNT}
\end{figure}
However, in processing the third clause (\theExmpl c), the reader may
recognize that it is the situation described {\em jointly} by
(12b) and (12c) that explains the situation described in (\theExmpl a). This
re-analysis of the relationship between the situations conveyed
by (12a) and (12b) correlates with the
tree restructuring operation of {\em adjunction to a leaf} -- in particular,
to the terminal on the {\em right frontier}. This second step corresponds
to step~2 of Figure~8. Notice how the {\em right frontier} of the tree
has changed, and thus which regions of the discourse model will be taken
to be in focus.

Finally, to get a feel for the discourse correlate of {\em adjunction to a
non-terminal node}, consider the following continuation of (\theExmpl).

\startx
a. John hates snakes.\\
b. His next-door neighbour had kept snakes,\\
c. and he had hated his neighbours.\\
d. Later in college, his roommate had kept snakes.
\stopx
Having decided that it is the situation described in (13 b,c) that provides
evidence for (13a), the reader in processing (13d) may decide that it is the
{\em combination} of situations (one described in 13b-c and the other
described in 13d) that provides evidence for (13a). This reanalysis of the
relationship between situations correlates with the tree restructuring
operation of {\em adjunction to a non-terminal}. This third step corresponds
to step~3 of Figure~8. As in step~2, the {\em right frontier} of the DST
has changed, and thus, the regions of the discourse model that are taken
to be {\em in focus}.

Note that because operations only apply to the right frontier of the DST,
the rest of the tree becomes irrelevant. In this light, one can easily
suppose that the only segments of the discourse and regions of the discourse
model that retain their identity are those that corresond to nodes on the
right frontier. When they no longer do so, the only thing that needs to
persist, for a listener to be said to have understood the discourse, are
the entities, propositions and relations predicated of them.

Finally, note that a similar assumption -- that integration of the
discourse meaning of the next clause only takes place at the
{\em right frontier} of the discourse structure --  is made by
Polanyi (1986) and Scha and Polanyi (1988).
There is also a close relationship between the notion of {\em right frontier}
presented here and Grosz and Sidner's (1986) stack of focus spaces
representing {\em Attentional State}. 

In Grosz and Sidner's (1986) model, a listener's attention at
any point correlates with the perceived structure of the discourse. In 
particular, they associate a {\em focus space\/} with each discourse 
segment, as well  as its {\em discourse segment purpose} (DSP). A focus space
contains discourse entities, along with the properties and relations predicated
of them. Corresponding to the evolving {\em Intentional Structure} (a tree of
DSPs), they propose a {\em stack} of focus spaces which represents the
listener's {\em Attentional State}. A segment's focus space is  pushed on the
stack when its DSP is taken to contribute to that of the  segment whose focus
space is at the top of the stack. Focus spaces  will be popped from the stack
prior to a push until the top focus space is one whose associated DSP can be
taken to {\em dominate} that associated with the focus space about to be
pushed. Another way to put this is that the stack contains the focus spaces of 
segments whose ``purposes'' are still open to additional support.

Drawing on Grosz's earlier work (Grosz, 1977, 1981), {\em Attentional State}
serves as a structured domain of locality for the interpretation  of definite
noun phrases. Resolution algorithms can then prefer to resolve a definite noun
phrase against a referent in a focus  space closer to the top of the stack than
in one further down. Picking up a referent further down may, in fact,
indicate that the segments associated with focus spaces higher up the stack
can now be taken as ``closed'', with attention shifting back to a more
inclusive segment.

The {\em right frontier} of the DST discussed here is closely related to
Grosz and Sidner's {\em Attentional State}. First, there is
a simple mechanical relationship: Any directed path through a tree from
root to leaf node can be mapped directly to a stack, with the element
corresponding to the leaf at the top. {\em Attach} and {\em
adjoin} then correspond exactly to the simple push and the sequence of pops
followed by a push that Grosz and Sidner use to manage the stack.
(That is, {\em attaching} a new leaf node corresponds to {\em pushing} a new
element on the stack. {\em Adjoining} a new node S$_{k}$ to a node
S$_{i}$ corresponds to {\em popping} all the stack elements through that
corresponding to S$_{i}$ and {\em pushing} that corresponding to
S$_{k}$ onto the top of the stack.)

There is also a functional relationship: nodes on the {\em right frontier} of
the DST correlate with regions of the discourse model taken to be
in focus -- substructures that resemble Grosz and Sidner's focus spaces.
In fact,
by positing a single tree structure and insertion algorithm to serve as a
formal analogue of both on-line recognition of discourse structure and changes
in participants' attention on regions of the discourse model, one can
eliminate Grosz and Sidner's stack as now redundant, while retaining their
insight into the usefulness of
distinguishing text structure, Intentional Structure and Attentional State.

\section{The Referents of Deictic  Pronouns}
\label{IDP}

The point of reviewing notions of discourse structure and 
incremental tree construction algorithms is to allow me to argue that it
is only regions of the discourse model corresponding to nodes on the
right frontier of the DST -- those regions that are {\em in focus} -- that
can yield referents for deictic pronouns. Before doing so, I want to
demonstrate that it is the structure of discourse segments (and hence
that of the discourse model) that constrains the referents of deictic
pronouns rather than the world being described.\footnote{I informally
analyzed 177 consecutive instances of pronominal 
reference using {\em it}, \this and {\em that}, distinguishing those 
that could be taken to co-refer with some noun phrase and those that 
could only possibly be taken to refer to the interpretation of one or 
more clauses. There were 96 instances of the latter. Of those, only 15 
($\sim$16\%) used the pronoun {\em it} while the other 81 
($\sim$84\%) used either \this or \that (19 instances of \that and 62 
instances of {\em this}). Of the 81 that co-referred with a noun 
phrase, 79 ($\sim$98\%) used {\em it} while only 2 ($\sim$2\%) 
used \this or {\em that}. My data comes from Taylor (1986); Hillis (1988);
an editorial from {\em The Guardian}, 15 December 1987; two reviews in
{\em TLS}, 23-29  October 1987, pp.1163-1164 and 20-26
November 1987, p.1270; and Kaelbling(1987).}

Evidence that the primary constraint on possible referents of the deictic
pronouns is the presentation of information (not simply what that information
is about) comes from the fact that, presented differently, the same information
about a situation gives rise to different referents. To see this, consider
(14), focussing on the referents of the deictic pronouns in b-d.

\startx

{\bf a.} There's two houses you might be interested in:

\vspace*{0.5ex}
{\bf b.} House A is in Palo Alto. It's got 3 bedrooms and 2 baths, and was
built in 1950. It's on a quarter acre, with a  lovely garden, and the owner is
asking \$425K. But \that's all I know about it.

\vspace*{0.5ex}
{\bf c.} House B is in Portola Vally. It's got 3 bedrooms, 4 baths and a
kidney-shaped pool, and was also built in 1950.  It's on 4 acres of steep
wooded slope, with a view of the mountains. The owner is asking \$600K. I
heard all \this from a real-estate friend of mine.

\vspace*{0.5ex}
{\bf d.} Is \that enough information for you to decide which to look at?
\stopx
\setcounter{HoldEx}{\value{Exmpl}}

That in each case it is an interpretation of an immediately preceding 
segment that, through its contribution to the discourse model, yields the
referent of the deictic pronoun, can be seen by presenting 
the same information in an interleaved fashion, a technique often 
used when comparing two items:

\startx

{\bf a.} There's two houses you might be interested in:

\vspace*{0.5ex}
{\bf b.} House A is in Palo Alto, House B in Portola Vally.  Both were built in
1950, and both have 3 bedrooms.  House A has 2 baths, and B, 4. House B also
has  a kidney-shaped pool.  House A is on a quarter acre, with a lovely
garden, while House B is on 4 acres of steep wooded slope, with a view of the
mountains. The owner of  House A is asking \$425K. The owner of House B is
asking \$600K. {\em That}'s all I know about House A. {\em This/That} I heard
from a real-estate friend of mine.
\vspace*{0.5ex}
{\bf 3.} Is \that enough information for you to decide which to look at?
\stopx
The two examples clearly have different segmental structures (at a gross level,
one that corresponds to the structure of the paragraphs). The question that
readers should ask themselves is whether the deictic pronouns in paragraphs
in (15b,c) have the same referents as they do in Example (14).
I believe they do not, and that this is because it is not the houses being
referred to, but rather a distinct chunk of information one has been told
about the houses.
Example \theExmpl\ does not contain separate segments describing what
the speaker knows about house A and about house B. Rather, there is only one
discourse segment containing information about both  houses. The only deictic
that refers easily and successfully is the final {\em that}, which successfully
refers to the information conveyed about both houses through entire segment.

Another piece of evidence that it is focused regions of the discourse
model that yield referents for deictic pronouns comes from the (partially)
recursive nature of discourse structure (cf. Section~\ref{DSegs}). At any
given point in a discourse, segments embedded at different depths can
yield referents for deictic pronouns. To see this, consider the following
quote from (Hillis, 1988):

\startx
\ldots it should be possible to identify certain functions as being
unnecessary for thought by studying patients  whose cognitive abilities are
unaffected by locally confined damage to the brain. $\{_{1}$For example,
binocular stereo fusion is known to take place in a specific area of
the cortex near the back of the head. $\{_{2}$Patients with damage to this
area of the cortex have visual handicaps but $\{_{3}$ [they] show no 
obvious
impairment in their ability to think.$_{3}\}_{2}\}$ {\em This\/}$_{i}$ 
suggests that
stereo fusion is not necessary for thought.$_{1}\}$ {\em This\/}$_{j}$ is a
simple example, and the conclusion is not surprising\ldots. 
(Hillis, 1988, p.185)
\stopx
\setcounter{SaveEx}{\value{Exmpl}}
(I have added brackets to indicate some of discourse segments at the points
where deictic pronouns occur, with subscripts indicating the depth of
embedding.) The most likely referent for {\em this}$_{i}$ is the
fact that visual cortex-damaged  patients have visual handicaps but no
impairment to their thinking abilities. This comes from Segment 2.
The most likely referent for {\em this}$_{j}$ is the whole ``brain 
damage'' example. This comes from the more inclusive Segment 1.

Not only do deictic pronouns get their referents from regions of the
model corresponding to nodes on the {\em right frontier} of the current
DST. These are the {\em only} regions that can provide such referents.
Consider the following variation of (\theHoldEx). (Here, the individual
clauses are  numbered for later discussion.)

\startx

{\bf a.} (1) There's two houses you might be interested in:

\vspace*{0.5ex}
{\bf b.} (2) House A is in Palo Alto. (3) It's got three bedrooms and two
baths, and was built in 1950. (4) It's on a quarter acre, with a lovely garden,
and (5) the owner is asking \$425K.

\vspace*{0.5ex}
{\bf c.} (6) House B is in Portola Vally. (7) It's got three bedrooms, four
baths  and a kidney-shaped pool, and (8) was also built in 1950.  (9) It's
on 4 acres of steep wooded slope, with a view of the mountains. (10) The owner 
is asking \$600K. (11) I heard all \this from a real-estate friend of mine.
(12)  But {\em that}'s all I know about House A.

\begin{figure}
\centerline{\psfig{figure=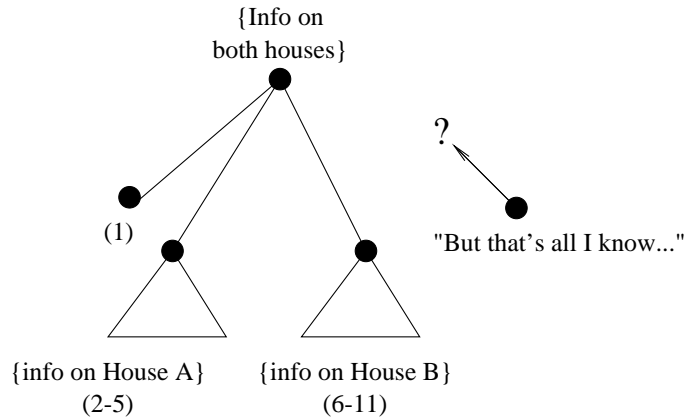}}
\caption{Discourse Segmentation at the point of processing
``But that's all \ldots''}
\label{pds-fig}
\end{figure}

\vspace*{0.5ex}
{\bf d.} (13) Is \that enough information for you to decide which to look at?
\stopx
What is at issue is the referent of (unstressed) \that in clause (12). The rest
of the clause constrains the referent of {\em that\/} to be 
information
about House A. However its position in the text is only compatible 
with its referring in one of three ways:
\begin{itemize}
\item It can co-refer with {\em all this} in clause (11), as in
``But {\em that}'s all she said.'' (In this paper, I do not discuss
deictic pronouns that refer to NP-evoked entities.)
\item It can refer to something associated with clause (11), such as
its corresponding assertion as in ``Of course,
{\em that}'s what I always tell people.''
\item It can refer to something associated with clauses 2-11 (the information
regarding both houses), similar to the perceived interpretation of
\that in clause 13.
\end{itemize}
Schematically, one might represent the discourse
segmentation at the point in processing \that, roughly as in Figure
\ref{pds-fig}. The oddity of
Example~\theExmpl\ comes from the conflicting demands of text 
position and clause predication in the process of resolving {\em that}.

Let me emphasize here that I am only considering written text and 
unstressed
instances of \this and {\em that}. It is well-known that stressing a 
pronoun
can shift its preferred referent. In the case of clause (12), stressing 
{\em that}, reinforced by information conveyed by the rest of the 
sentence, can force its referent to be the block of information about
House A, even though its corresponding region in the discourse model
is no longer {\em in focus}.

Notice that even if it is true that unstressed \this and \that must come
from focused regions of the model (ones corresonding to nodes on the right
frontier of the DST), there is still an ambiguity as to {\em which} region.
To see this, consider the first part
of the Hillis' example (repeated here) as a ``discourse completion task''.

\startx

\ldots it should be possible to identify certain functions as being
unnecessary for thought by studying patients  whose cognitive abilities are
unaffected by locally confined damage to the brain. For example, binocular
stereo  fusion is known to take place in a specific area of the cortex near
the back of the head. Patients with damage to  this area of the cortex have
visual handicaps but show no obvious impairment in their ability to think.
{\em This} \ldots.
\stopx
At this point in the discourse, there are many possible ways of 
completing
the last sentence, among them --

\startx

a. {\em This} is obvious when they are asked to solve word problems
presented orally.\\
b. {\em This} suggests that stereo fusion is not necessary for thought.\\
c. {\em This} is only a simple example, and the conclusion is
not surprising.
\stopx
In (19a), \this refers to the {\em fact} that patients with
damage to the area of the cortex
near the back of the head show no obvious impairment in their 
ability to think. In (19b), \this must refer to the more
inclusive claim that patients with damage to the particular area of 
the cortex near the back of the head have visual handicaps but show 
no obvious impairment in their ability to think. (This is what shows 
that one doesn't need stereo fusion in order to think.) Finally, in (19c)
\this clearly refers to the entire example about binocular stereo vision.
Which discourse segment provides the referent
for \this depends on what is compatible  with the meaning of the rest
of the sentence, as I noted in Section~\ref{terminology} in connection with
example (20):

\startx
{\bf a.} Segal, however, had his own problems with women: he had been trying
to keep his marriage of seven years from falling apart; when {\em that} became
impossible \ldots
\begin{center}
that $\equiv$ keeping his marriage from falling apart
\end{center}
\vspace*{2ex}
{\bf b.} Segal, however, had his own problems with women: he had been trying
to keep his marriage of seven years from falling apart; when {\em that} became
inevitable \ldots
\begin{center}
that $\equiv$ his marriage falling apart
\end{center}
\stopx
As with other types of ambiguity, there may be a default
preference (e.g., based on recency, position, etc.) in a ``neutral'' context
but, if there is one, it can easily be over-ridden by the demands of the
actual context (Crain \& Steedman, 1985; Steedman, 1989).

This ambiguity as to where a deictic pronoun takes its referent seems very
similar to the ambiguity associated  with the use of deixis for pointing within
a shared physical context. Both Quine (1971) and Miller (1982) have
observed in this regard that {\em all pointing} is ambiguous: the intended
demonstratum of a pointing gesture can be any of the infinite number of
points "intersected" by the gesture or any of the structures encompassing
those points. (Or, one might add, any interpretation of those 
structures.) The ambiguity here -- how inclusive a segment is
providing an interpretation for \this or \that -- seems very similar.

\section{Referring Functions and the Interpretation of Deictic Pronouns}
\label{Bronto}

Having set up what I hope is all the necessary groundwork in the previous
three sections, I hope now to pull it together and show how deictic pronouns
get their referents. The solution I propose is based on distinguishing
between what can be {\em pointed to} (the {\em demonstratum}) and what
can be {\em referred to} by virtue of pointing (the {\em referent}). I
claim that it is focussed regions of the discourse model that deictic
pronouns can point to (via the discourse entity ``proxies'' of those
segments), and that it is possibly new discourse entities that they can refer
to by virtue of that pointing. What links the two, I claim, are what
Nunberg (1979) has called {\em referring functions}.

Nunberg (1979) introduces the notion of referring function in connection
with some rather extreme examples of demonstratum/referent pairs,
such as in (17) below (Nunberg's numbering):
\begin{quote}
For example, a restaurant waiter going off duty might remind his replacement:

(16) The ham sandwich is sitting at table 20.

And in just those constexts, he could equally well point at a ham sandwich and
say,

(17) He is sitting at table 20. (Nunberg 1979, p.149)
\end{quote}
Here, the ham sandwich serves as the demonstratum of the pointing act
and the guy who ordered the ham sandwich serves as the referent of the
associated pronoun ``He''. Formally, {\em referring functions} map
{\em demonstrata} into {\em intended referents}.
\begin{center}
{\bf f}: {\bf D} $\rightarrow$ {\bf R}
\end{center}
where {\bf f} is the referring function, {\bf D} is the domain of
demonstrata and {\bf R}, the range of intended referents. The problem for
the listener, given the ambiguity of pointing mentioned in
Section~\ref{IDP}, is to simultaneously constrain the demonstratum
and the referring function, such that the intended referent results.

Aiding the listener are factors that Nunberg takes as constraining the
referring functions that can be used in a particular case. These include:
\begin{itemize}
\item that {\bf R}, the range of the referring function, intersect ``the
set of things that the speaker might rationally be construed as intending
to refer to in a given context'', where
the latter is determined by ``the nature of the predication, by the morphology
of the demonstrative pronoun, and by such contextual considerations at `topic
of conversation''' (Nunberg, 1979, p.157);
\item that it be possible for the
listener to determine, in practice, just what stands in that relationship.
(For example, if one wants to refer to a particular {\em place} by pointing
to a particular object, the listener should recognize what place is salient
to that object. A wine bottle labelled ``Medoc'' may make the Medoc a
salient place, while a bottle of Coca Cola may not make any particular
place salient.)
\item that it be a more likely referring function, under the
circumstance, than any other.
\end{itemize}
Similar assumptions are made by Hirschberg (1986) with respect to which
{\em scales} a spekaer can use to make scalar implications in a given case.

To see that these assumptions make sense, consider the ``ham sandwich''
example given earlier. Here, {\bf R} must intersect with males (given
the pronoun ``he'') who are able to sit at a table (given the verb phrase).
The listener can be expected to figure out which particular person stands
in that relation, because it must be a man sitting at table 20. Finally,
the circumstances of an order waiting in the kitchen to be delivered
make it more like that {\em f} picks out ``the man who ordered the
hame sandwich'' than ``the man who made it''.

I claim that the same approach can be extended to cover
discourse deixis as well as external ostension, simply by taking a different
domain {\bf D} of demonstrata. Specifically, I noted in
Section~\ref{terminology} that I was making the same assumption as
that made in DRT (Asher, 1987; B\"auerle, 1989) that each context --
region of the discourse model -- has a discourse entity that ``stands
proxy'' for its propositional content. If one takes {\bf D} to comprise
the discourse entity ``proxies'' of those regions of the discourse model
that are currently {\em in focus} -- those corresponding to (and whose
discourse segments correspond to) nodes on the {\em right frontier} of the
DST, then everything can proceed as before.

To illustrate this approach, I will apply it to the first examples given
in the paper (repeated here, with the initial clauses numbered):

\startx
It's always been presumed that (2) when the glaciers receded, (3) the 
area got very hot. (4) The Folsum men couldn't adapt, and (5) they
died out. (6) {\em That}'s what is supposed to have happened. It's the
textbook dogma. But it's  wrong.
\stopx
\startx
(1) Using microscopes and lasers and ultrasound, (2) he removes tumors
(3) that are intertwined with children's brain stems and spinal cords.
(4) There is only the most minute visual difference between the tumors and
normal tissue. (5) Operations can last 12 hours or more. (6) The tiniest
slip can kill, paralyze or leave a child mentally retarded.

(7) {\em This} is the easy part of his job. \hspace*{2em}({\em New York Times},
11 August 1990)
\stopx

First consider example (21). Here at least four regions of the discourse
model can be said to be {\em in focus} at the point in the discourse at
which {\em that} appears:
\begin{itemize}
\addtolength{\itemsep}{-8pt}
\item the region associated with clause 5,
\item the region associated with clauses 4 and 5,
\item the region associated with clauses 2-5, and
\item the region associated with clauses 1-5.
\end{itemize}
Thus {\bf D}, the domain of {\bf f}, will comprise at least their four
``proxies''. {\bf R}, the range of {\bf f}, should be a subset of event
tokens (i.e things that can happen). It is easy to imagine a function
that could apply to each of the first three proxies to yield an event
token, because their propositional content can be taken to convey a
particular event. The intended referent could be any of these, but given
the rest of the story (Tony Hillerman's {\em Dance Hall of the Dead}), it
is probably either the second or third.

Now consider example (22). Here at least two regions of the discourse model
can be said to be {\em in focus}:
\begin{itemize}
\addtolength{\itemsep}{-8pt}
\item the region associated with clause 6, and
\item the region associated with clauses 1-6.
\end{itemize}
Thus {\bf D} will comprise at least their two proxies. I assume that it
is intentional act types that are parts of jobs, so that {\bf R} will
intersect this set. It is hard to associate an intentional act type with
the first proxy. (If clause 7 had been something like ``{\em This} would
paralyze some people into inaction'', where {\bf R} would comprise
(probably difficult) situations, it would be easier to associate such
a referent with the first proxy.) So the intended referent of {\em that}
in cluase 7 most likely derives from applying {\bf f} to the second
proxy -- most likely paraphrasable as ``removing these difficult tumors''.

Notice that this approach in terms of referring functions avoids a problem
that B\"{a}uerle (1989) cannot, because his approach requires separate
mechanisms to handle event token reference, event type reference and
proposition (token) reference. That is, the current approach does not
require introducing yet a fourth mechanism to handle deictic reference to
proposition types (not discussed in B\"{a}uerle's paper) such as the
following:

\startx
a. Fred and Amy always cheat on their homework.\\
b. Well, I'd believe {\em that} of Fred, but not of Amy.
\stopx

Finally, let me stress both the similarities and differences between
this account of discourse deixis and the process called
{\em accommodation} (Section~\ref{terminology}). Like {\em accommodation},
this account of discourse deixis postulates that new individuals --
new entities in the discourse model -- can be introduced by virtue of a
referring action. These entities have the same status in the model as
any introduced through the use of an indefinite noun phrase. That event
tokens, event types, facts, descriptions, actions, etc., introduced
this way can be individuals in their own right seems in line with the
view gaining currency in formal semantics that the domain of the type
{\em individual} has a very rich substructure (Bach, 1989; Link, 1983;
1987; Schubert \& Pelletier, 1987).

However, one should maintain a distinction between discourse deixis and
the more general process of {\em accommodation}, because the former is
more constrained. In accommodating a definite noun phrase, the discourse
context must provide the listener with a discourse entity with which
he or she can presume to associate a unique individual that satisfies
the descriptive content of the noun phrase. Understanding definite
noun phrases can thus be said to require all of a listener's world
knowledge. Unlike a definite noun phrase, however, a deictic pronoun
has no descriptive content of its own. Thus what the discourse context
must provide is a small set of entities that can be pointed to, whose
propositional content is the source of new intended referents.
Finding an appropriate referrring function and intended referent in
this case seems to be a less comprehensive task.

\section{Conclusion}

In this paper I have argued for an account of discourse deixis, based on
current views of discourse structure. In particular, I argued that what
provides referents for these expressions are the interpretations of
discourse segments corresponding to nodes on the {\em right frontier} of
a formal tree structure analogue. I also argued that the manner by which
these expressions get their
referents could be viewed as an extension of Nunberg's use of referring
functions. For natural-language understanding systems, what lies ahead
now is more work on discourse-level semantics, so that (1) systems
can build appropriately structured discourse models in response to
a text and (2) they can identify intended demonstrata and reference
functions and thereby resolve this interesting class of referring forms.


\begin{thebibliography}{99}

\bibitem{alle87} Allen, J. {\em Natural Language Understanding}.
Menlo Park CA: Benjamin/Cummings Publ. Co., 1987.

\bibitem{asher}
Asher, N. A Typology for Attitude Verbs and their Anaphoric Properties.
{\em Linguistics and Philosophy} 10, pp. 125-197, 1987.

\bibitem{asher2}
Asher, N. {\em Abstract objects, semantics and anaphora}. (Forthcoming) 

\bibitem{bach} Bach, E. {\em Informal Lectures on Formal Semantics}.
Albany NY: State University of NY Press, 1989.

\bibitem{bauer88} B\"{a}uerle, R. Aspects of Anaphoric Reference to Events
and Propositions in German. In C. Rohrer and Wedekind (eds.){\em untitled}.
Reidel. To appear. Also appears as Ev\'{e}nements et propositions: quelques
aspects de la r\'{e}f\'{e}rence anaphorique en allemand. In P. Engel (ed.),
{\em
Recherches sur la philosophie et le langage 10} (Cahier du groupe de recherches
sur la philosophie et le langage, Universit\'{e} des Sciences Sociales de
Grenoble), 1989.

\bibitem{cohe83} Cohen, R. {\em A Computational Model for the Analysis
of Arguments}. PhD thesis, Technical Report 151, Computer Systems
Research Group, University of Toronto, 1983.

\bibitem{cohe87} Cohen, R. Analyzing the Structure of Argumentative
Discourse,  {\em Computational Linguistics} 13(1-2):11-24, 1987.

\bibitem{crai85} Crain, S. and Steedman, M. On not being led up the 
garden path:
the use of context by the psychological  parser. In {\em Natural Language 
Parsing}, D. Dowty, L. Karttunen \& A. Zwicky (eds.), Cambridge: 
Cambridge University Press, 1985.

\bibitem{dieu89} Di Eugenio, B. Clausal Reference in Italian. {\em Proc. Penn
Linguistics Conference}, University of Pennsylvania, February 1989.

\bibitem{fox87} Fox, B. {\em Discourse Structure and Anaphora}. 
Cambridge: Cambridge University Press, 1987.

\bibitem{gros77} Grosz, B. The Representation and Use of Focus in a System
for Dialog Understanding.  Technical Report 151, SRI International, Menlo
Park CA, 1977.

\bibitem{gros81} Grosz, B. The Representation and Use of Focus in a System
for Understanding Dialogs. In {\em Elements of  Discourse Understanding},
A. Joshi, B. Webber and I. Sag (eds.), Cambridge: Cambridge Univ. Press, 
1981. (Reprinted in {\em Readings in Natural Language Processing}, B. 
Grosz, K. Sparck Jones and B. Webber (eds.),  Los Altos: Morgan Kaufmann 
Publ., 1986.)

\bibitem{gjw83} Grosz, B., Joshi, A. and Weinstein, S. Providing a
Unified Account of Noun Phrases in Discourse. {\em Proceedings of the
21st Annual Meeting of the Association for Computational Linguistics},
pp.44-50.

\bibitem{gros86} Grosz, B. and Sidner, C. Attention, Intention and 
the Structure
of Discourse. {\em Computational Linguistics},  12(3), July-Sept. 1986,
pp.175-204.

\bibitem{heim83} Heim, I. File Change Semantics and the Familiarity Theory
of Definiteness. In {\em Meaning, Use and Interpretation of Language}, R.
Bauerle, C. Schwarze and A. von Stechow (eds.). Berlin: de Gruyter, 1983.

\bibitem{hill88} Hillis, W.D. Intelligence as an Emergent Behavior,
{\em Daedalus}, Winter 1988, pp.175-190.

\bibitem{hind79} Hinds, J. Organizational Patterns in Discourse.
In T. Givon (ed.) {\em Syntax and Semantics 12: Discourse and
Syntax}, New York: Academic Press, 1979.

\bibitem{hirs86} Hirschberg, J. {\em A Theory of Scalar Implicature}.
PhD thesis, Department of Computer \& Information Science, University
of Pennsylvania, Philadelphia PA, 1986.

\bibitem{hirs87} Hirschberg, J. \& Litman, D. Now Let's Talk about
Now: Identifying Cue Phrases Intonationally. {\em Proc. 25th  Annual 
Meeting,
Association for Computational Linguistics}, Stanford Univ. Stanford 
CA, July 1987.

\bibitem{hobb88} Hobbs, J. , Stickel, M., Martin, P. \& Edwards, D.
Interpretation as Abduction. {\em Proc. 26th Annual  Meeting, 
Association for
Computational Linguistics}, SUNY Buffalo, Buffalo NY, June 1988, 
pp.95-103.

\bibitem{kael87} Kaelbling, L.  An Architecture for Intelligent Reactive
Systems. Technical report, SRI International, Menlo  Park CA, 1987.

\bibitem{kamp81} Kamp, H. A Theory of Truth and Semantic Representation.
In {\em Formal Methods in the Study of Language}, J. Groenendijk, T.
Janssen and M. Stokhof (eds). Amsterdam: Mathematical Center Tracts, 1981.

\bibitem{kart76} Karttunen, L. Discourse Referents. In {\em Syntax and
Semantics}, Volume 7, J. McCawley (ed.). New York: Academic Press, 1976.

\bibitem{lakoff} Lakoff, R. Remarks on {\em this} and {\em that}. {\em
Proceedings of the Chicago Linguistics Society}, 1974, pp.345-356.

\bibitem{lewis79}
Lewis, D. Score Keeping in a Language Game. In R. B\"{a}uerle et al. (eds.),
{\em Semantics from a Different Point of View}. Berlin: Springer Verlag, 1979.

\bibitem{lind79} Linde, C. Focus of Attention and the Choice of Pronouns in
Discourse. In T. Givon (ed.) {\em Syntax and Semantics 12: Discourse and
Syntax}, New York: Academic Press, 1979.

\bibitem{link83} Link, G. The Logical Analysis of Plurals and Mass Terms:
A lattice-theoretical approach. In R. B\"{a}uerle {\em et al.} {\em
Meaning, Use and the Interpretation of Language}, Berlin: de Gruyter, 1983,
pp.302-323.

\bibitem{link87} Link, G. Algebraic Semantics for Event Structures. In
Groenendijk, Stokhof and Veltman (eds.), {\em Proc. 6th Amsterdam Colloquium
ITLI}, Amsterdam, 1987.

\bibitem{lyons} Lyons, J. {\em Semantics}. Cambridge: Cambridge University
Press, 1977.

\bibitem{mill82} Miller, G. Problems in the Theory of Demonstrative Reference.
In R. Jarvella \& W. Klein (eds.), {\em Speech, Place and Action}.  
New York: Wiley, 1982.

\bibitem{nakh88} Nakhimovsky, A. Aspect, Aspectual Class and the Temporal
Structure of Narrative. {\em Computational  Linguistics}, 14(2):29-43, 1988.

\bibitem{nunb79}
Nunberg, G. The Non-Uniqueness of Semantic Solutions: Polysemy. {\em Lingustics
and Philosophy}, 3(2):143-184, 1979.

\bibitem{pass89} Passonneau, R. Getting at Discourse Referents,
{\em Proc. 27th Annual Meeting, Association
for Computational Linguistics}, Univ. British Colubia, Vancouver, Canada,
June 1989.

\bibitem{pola86} Polanyi, L. The Linguistic Discourse Model: Towards 
a formal
theory of discourse structure. TR-6409. BBN  Laboratories Incorp., 
Cambridge
MA, November 1986.

\bibitem{prin86} Prince, E. On the Syntactic Marking of Presupposed Open
Propositions. {\em Papers from the Parasession on Pragmatics and Linguistic
Theory}. 22nd Annual Regional Meeting of the Chicago Linguistics Society.
Chicago IL, 1986, pp. 202-222.

\bibitem{quin71} Quine, W. The Inscrutability of Reference. In D. 
Steinberg
and L. Jacobovits (eds.), {\em Semantics: An Interdisciplinary 
Reader}. 
Cambridge: Cambridge University Press, 1971. pp.142-154.

\bibitem{reich} Reichman, R. {\em Getting Computers to Talk like You 
and Me}.
Cambridge MA: MIT Press, 1985.

\bibitem{rein83}
Reingold, E. and Hansen, W. {\em Data Structures}. Boston: Little, Brown and
Co., 1983.

\bibitem{robe89}
Roberts, C. Modal Subordination and Pronominal Anaphora in Discourse.
{\em Linguistics and Philosophy} 12(6), pp. 683-721, 1989.

\bibitem{sadock} Sadock, J. Read at Your Own Risk: Syntactic and semantic
horrors you can find in your medicine chest. {\em Proc. Chicago Linguistics
Society} 10, 1974, pp.599-607.

\bibitem{scha88} Scha, R. and Polanyi, L. An Augmented Context Free
Grammar for Discourse. {\em Proceedings of the 12th International
Conference on Computational Linguistics}, August 1988, Budapest
Hungary.

\bibitem{schu87} Schubert, L., and Pelletier, F. Problems in the
Representation of the Logical Form of Generics, Plurals, and Mass
Nouns. In {\em New Directions in Semantics}. London: Academic Press, 1987,
pp.385-451.

\bibitem{sidn83} Sidner, C. Focusing in the Comprehension of
Definite Anaphora. In M. Brady \& R. Berwick (eds.), {\em
Computational Models of  Discourse}.
Cambridge MA: MIT Press, 1982, pp.267-330.

\bibitem{stee89} Steedman, M. Grammar, Interpretation and Processing from
the Lexicon. In W. Marslen-Wilson (ed.) {\em Lexical Representation and
Process}. Cambridge MA: MIT Press (Bradford Books), 1989.

\bibitem{stee90} Steedman, M. Structure and Intonation in Spoken Language
Understanding. {\em Proceedings of the 28th Annual Meeting of the Association
for Computation Linguistics}, Pittsburgh PA, 1990.

\bibitem{strunk} Strunk, W. Jr. and White, E.B. {\em Elements of Style}
third edition), New York: Macmillan Co, 1959.

\bibitem{tayl} Taylor, P. {\em Summons to Memphis}. New York: Ballentine
Books, 1986.

\bibitem{webb79}
Webber, B. {\em A Formal Approach to Discourse Anaphora}. New York: Garland
Press, 1979.

\bibitem{webb83} Webber, B. So What can we Talk about Now? In M. Brady and
R. Berwick (eds.), {\em Computational Models of Discourse}. Cambridge MA: MIT
Press, 1982, pp.331-371.

\bibitem{webb88} Webber, B. Tense as Definite Anaphor. {\em Computational
Linguistics}, 14(2):61-73, 1988.

\bibitem{ws79} Wilson, D. and Sperber, D. Ordered Entailments. In {\em
Syntax and Semantics XI: Presuppositions}. London: Academic Press, 1979.
\end{thebibliography}
\end{document}